\documentclass{article}

\usepackage{microtype}
\usepackage{graphicx}
\usepackage{subfigure}
\usepackage{booktabs} 

\usepackage{hyperref}

\usepackage[accepted]{icml2024}

\usepackage{amsmath}
\usepackage{amssymb}
\usepackage{mathtools}
\usepackage{amsthm}
\usepackage{hyperref}

\usepackage{comment}
\usepackage{multicol}

\usepackage[capitalize,noabbrev]{cleveref}

\theoremstyle{plain}

\theoremstyle{definition}

\theoremstyle{remark}

\usepackage[textsize=tiny]{todonotes}

\icmltitlerunning{LeaPformer: Enabling Linear Transformers via Learned Proportions}

\begin{document}

\twocolumn[
\icmltitle{LeaPformer: Enabling Linear Transformers for Autoregressive and \\
Simultaneous Tasks via Learned Proportions}



\icmlsetsymbol{equal}{*}

\begin{icmlauthorlist}
\icmlauthor{Victor Agostinelli}{osu}
\icmlauthor{Sanghyun Hong}{osu}
\icmlauthor{Lizhong Chen}{osu}
\end{icmlauthorlist}

\icmlaffiliation{osu}{Oregon State University, OR USA}

\icmlcorrespondingauthor{Lizhong Chen}{chenliz@oregonstate.edu}

\icmlkeywords{Machine Learning, ICML, Linear Attention, Efficiency, Efficient Modeling, Transformers}

\vskip 0.3in
]



\printAffiliationsAndNotice{}  


\begin{abstract}

%
A promising approach to preserving model performance in linearized transformers is to employ position-based re-weighting functions. However, state-of-the-art re-weighting functions rely heavily on target sequence lengths, making it difficult or impossible to apply them to autoregressive and simultaneous tasks, where the target and sometimes even the input sequence length are unknown. To address this issue, we propose \textit{Learned Proportions} (LeaP) and LeaPformers\footnote[2]{\url{https://github.com/OSU-STARLAB/LeaPformer}}. Our contribution is built on two major components. First, we generalize the dependence on explicit positional representations and sequence lengths into dependence on sequence proportions for re-weighting. Second, we replace static positional representations with dynamic proportions derived via a compact module, enabling more flexible attention concentration patterns. We evaluate LeaPformer against eight representative efficient transformers on the Long-Range Arena benchmark, 
showing that LeaPformer achieves the best quality-throughput trade-off, as well as LeaPformer to Wikitext-103 
autoregressive language modeling and simultaneous speech-to-text translation for two language pairs, achieving competitive results. 


\end{abstract}

\section{Introduction}
\label{sec:intro}

Transformers \cite{attention_is_all_you_need} are dominant in the natural language processing (NLP) solution space
, demonstrating state-of-the-art performance for a range of applications. With the advent of widely accessible large language models (LLMs), transformers as a class of models are being studied more closely than ever. Unfortunately, the quadratic complexity of the attention mechanisms of typical transformers limits the lengths of the sequences that they can process, rendering them sub-optimal or even impossible to apply for tasks with long sequences. 

\begin{figure}[t]
    \centering
    \includegraphics[scale=0.53]{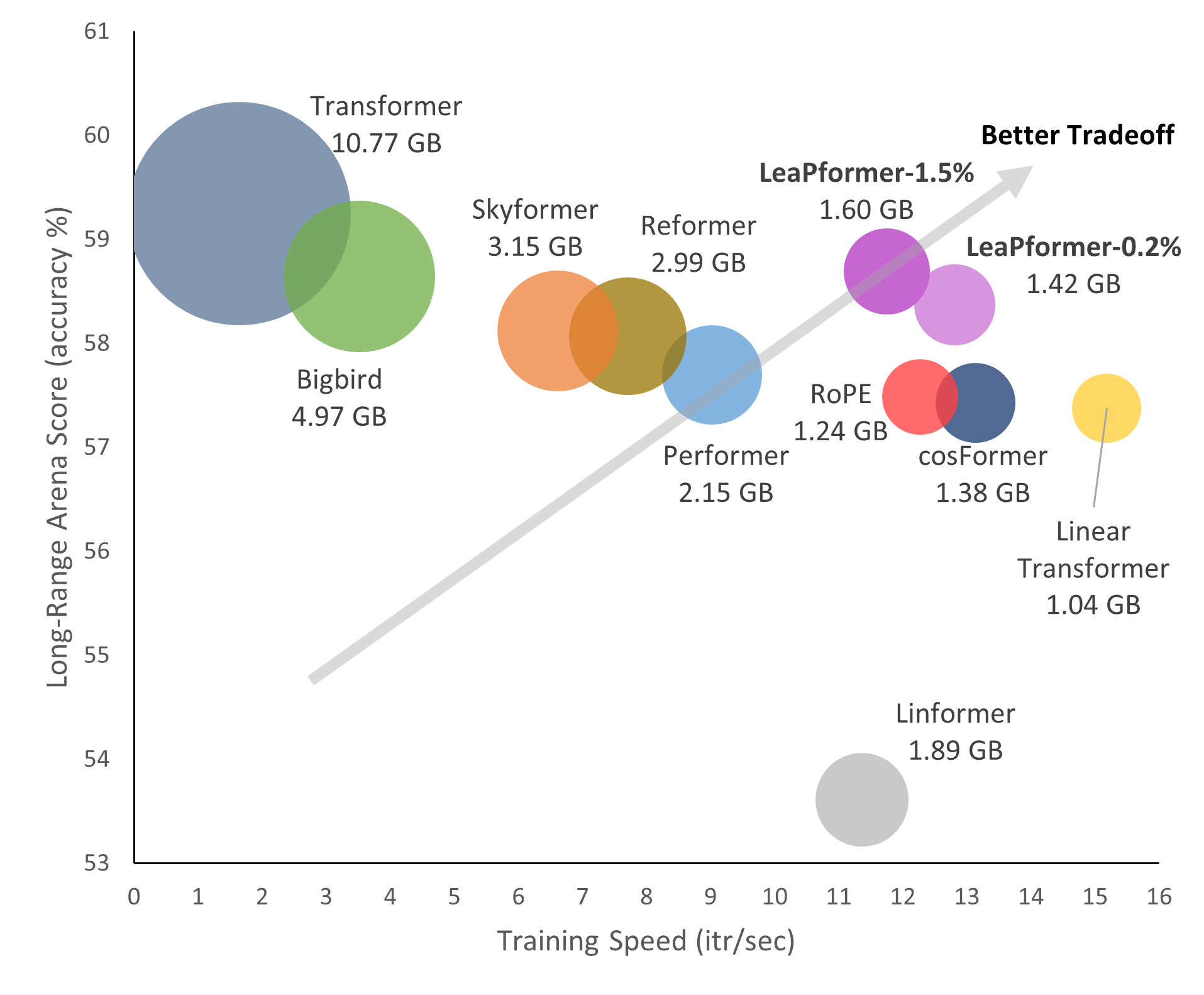}
    \vspace{-0.5em}
    \caption{\textbf{Contrasting accuracy-throughput trade-offs between LeaPformers and baselines.} Throughput for 4k sequence length tasks (x-axis) vs. average accuracy score (y-axis) across the five tasks in the Long-Range Arena benchmark. The memory footprint of each mechanisim is specified in labels and represented by circle size. \textbf{LeaPformers} provide the best average inference accuracy-throughput trade-off while achieving the second best overall score with only up to a 1.5\% increase to parameter count.}
    \label{fig:trade-off}
    \vspace{-1.5em}
\end{figure}

Naturally, an active area of possible improvement for classical transformers are efficient attention mechanisms that reduce the sometimes prohibitive quadratic run-time and memory complexity of softmax attention with respect to sequence lengths. Many efficient transformer variants have been proposed, including both sub-quadratic attention mechanisms, usually with key assumptions or experimental bounds surrounding their construction, and truly linear attention mechanisms with no prior environmental assumptions \cite{kathrapalous20, performer2020, rfapeng2021, chen2021skyformer, zhen2022cosformer}. While the aforementioned linear transformers are often effective for specific tasks, they tend to exhibit varying degrees of performance degradation when generalized.



To address this issue, re-weighting functions have been recently formalized \cite{su2022roformer, zhen2022cosformer} in linear transformers and serve to concentrate attention scores. Although promising, state-of-the-art position-based re-weighting functions rely on explicit token positions and sequence lengths \cite{zhen2022cosformer}. This reliance on knowing the sequence length beforehand make it difficult to apply those re-weighting functions and linear transformers to autoregressive tasks without specialized solutions \cite{agostinelli2023improving} and renders it impossible to apply them to simultaneous tasks. Furthermore, existing re-weighting functions' reliance on explicit positional representations usually produce static attention concentration patterns, which can severely limit their generalizability when an attention concentration pattern is ill-suited to a given task.


To solve this reliance on explicit positional representations and enable linear transformers for a wider range of tasks, we propose a novel approach that we refer to as \textit{Learned Proportions} (LeaP) and call models we apply it to LeaPformers. This contribution is composed of two major aspects: generalization to proportions and learned behavior. First, we generalize the dependence on explicit positional representations and sequence lengths into an intuitive dependence on proportions of a sequence for re-weighting, removing theoretical dependence on sequence lengths. Second, instead of employing static positional representations, we construct and deploy a compact module that dynamically derives sequence proportions for a given token during training and inference. These straightforward, but critical, contributions ultimately remove any reliance that current position-based re-weighting functions may have on sequence length, enabling them for tasks where the sequence length is not known beforehand (and cannot be estimated) and/or where attention concentration patterns are more complex.


To validate our proposed approach, we primarily test 
LeaPformer against cosFormer, the state-of-the-art position-based linear transformer, by adapting cosFormer's cosine-based re-weighting function via LeaP. We also evaluate and compare with eight other representative attention mechanisms on the Long-Range Arena (LRA) benchmark \cite{tay2021long}, a competitive benchmark for efficient attention mechanisms on long sequences. In addition, we validate LeaPformers on autoregressive language modeling on Wikitext-103b \citep{merity2016pointer} and on multiple language pairs for simultaneous speech-to-text translation (SimulST) \cite{simulst_xma20}. When compared to popular, previously proposed efficient attention mechanisms on the LRA benchmark, the proposed LeaPformer achieves the best accuracy-throughput trade-off, balanced performance across tasks, small memory footprint, and notably beats cosFormer's inference quality (see Figure \ref{fig:trade-off}). During autoregressive language modeling, LeaPformer achieves the lowest perplexity out of a limited set of efficient attention mechanisms, beating out the next closest mechanism by 0.13 perplexity on the test set. Finally, when applied to simultaneous translation, LeaPformer demonstrates competitive results with a reasonable accuracy-throughput trade-off compared to classical softmax attention for critical ablations, with 
variations achieving quality loss of only 0.26 BLEU-4 \cite{sacrebleu2018} for English to German and 0.23 BLEU-4 for French to English while being completely linear in complexity. To our knowledge, this is the first time that an explicit position-based re-weighting function for linear transformers is successfully applied to simultaneous tasks.



\section{Background and Motivation}
\label{sec:background}

Here we provide an overview of the background knowledge required to understand our work and motivate LeaPformers.

\subsection{Softmax Attention Mechanisms}
Multi-headed self-attention in transformers~\cite{attention_is_all_you_need} can generally be described as follows:

\vspace{-1em}

\begin{equation}
    \vspace{-2em}
    a_h(x) = softmax(\frac{Q_h K^T_h}{\sqrt{d}})V_h
    \label{eq:attn}
\end{equation}

\begin{equation}
    A(x) = concat(a_1(x), a_2(x), \; \dots \; , a_H(x))W_{out}
    \label{eq:concat}
\end{equation}
where query is $Q_h = xW_{q,h}$, key $K_h = xW_{k,h}$, and value $V_h = xW_{v,h}$, with $x \in \mathbb{R}^{n \times d_{model}}$ being the input sequence for each attention head that divides the model embedding space $d_{model}$ into some $d_{head}$ (denoted as $d$ hereafter for simplicity) and $W_{q, h} \in \mathbb{R}^{d_{model} \times d}$, $W_{k, h} \in \mathbb{R}^{d_{model} \times d}$ and $W_{v, h} \in \mathbb{R}^{d_{model} \times d}$. In cases where the concatenation of the attention head outputs differs in dimensionality from $d_{model}$, an optional output projection layer is commonly applied via $W_{out} \in \mathbb{R}^{d_{out} \times d_{model}}$.
%
For long sequences, the quadratic complexity of the mechanism 
in Equation \ref{eq:attn} can prove to be a throughput bottleneck 
during training and inference.

\subsection{Efficient and Linear Transformers}
Efficient 
and linear transformers have emerged over the past few years as an active area of research for particularly resource or latency-constrained environments, exhibiting notable inference speedups and smaller memory footprints. These transformer variants focus on alternative attention mechanisms that reduce the quadratic complexity of typical softmax attention. A plethora of efficient transformer options exist that can be classified into a few groups: sliding-window or localized attention mechanisms \cite{image_transf, transformerxl, memformer, longformer2020}, pattern or sparsity-based attention mechanisms \cite{sparsetransformer, bigbird2020}, kernel-based and truly linear attention mechanisms with no priors \cite{kathrapalous20, performer2020, rfapeng2021, chen2021skyformer, zhen2022cosformer}, and some unique outliers \cite{linformer2020, reformer2020}.

While many approaches linearize the computations, \emph{truly-linear} transformers, such as the kernel-based substitutions for the softmax mechanism, do not make any prior assumptions of the environments (e.g., no assumed sparsity or local dependencies).
This can be described 
via row-wise outputs (represented by $a_{h, i}(x)$) for each attention head in Equations \ref{eq:redistr_attn}, \ref{eq:split}, and \ref{eq:lin_attn}, with $S$ corresponding to any similarity function that transforms the product of the query and key matrices. 
If $S$ becomes $exp$, Equation \ref{eq:redistr_attn} is an accurate representation of softmax attention. 
If we decompose $S$ into $S_q$ and $S_k$, 
as shown in Equation \ref{eq:split}, computation can be reordered such that the attention complexity 
reduces from $O(N_1N_2d)$ in Equation \ref{eq:redistr_attn} 
to $O(N_1d^2 + N_2d^2)$ in Equation \ref{eq:lin_attn}. 
$N_1$ corresponds to the sequence length of the query matrix and $N_2$ corresponds to 
those of the key and value matrices (a generalization for encoder-decoder cross-attention). When $N_1$ or $N_2$ are significantly larger than $d$, this rearrangement of the attention calculation leads to linear complexity with respect to the sequence length. 



\begin{equation}
\vspace{-1.12em}
a_{h,i}(x) = \sum_j \frac{S(Q_{h,i}K_{h,j}^T)}{\sum_jS(Q_{h,i}K_{h,j}^T)}V_{h,j}
\label{eq:redistr_attn}
\end{equation}

\begin{equation}
\vspace{-1.12em}
decompose(S(Q_{h,i}K_{h,j}^T)) = S_q(Q_{h,i})S_k(K_{h,j}^T)
\label{eq:split}
\end{equation}

\begin{equation}
a_{h,i}(x) = \sum_j \frac{S_q(Q_{h,i})(S_k(K_{h,j}^T)V_{h,j})}{S_q(Q_{h,i})\sum_jS_k(K_{h,j}^T)}
\label{eq:lin_attn}
\end{equation}

\subsection{Position-Based Re-weighting Functions}
While reducing the computational complexity, linearizing multi-headed self-attention 
leads to varying degrees of degraded model performance. To address this shortcoming, re-weighting functions have been recently proposed. 
They introduce an additional function to augment $S(Q_{h,i}, K^T_{h,j})$, with the 
goal of concentrating/adjusting the probability distribution of the normalized $Q_hK_h^T$ \cite{zhen2022cosformer}. 
Re-weighting functions $\sigma(i, j)$ are commonly based on token positions, and we multiply as shown in Equation \ref{eq:gen_reweight}:

\vspace{-0.5em}

\begin{equation}
S(Q_{h,i}, K_{h,j}^T) = S_q(Q_{h,i})S_k(K_{h,j}^T)\sigma(i, j)
\label{eq:gen_reweight}
\end{equation}

Note that even though $\sigma(i, j)$ 
is placed at the end of the equation and multiplied, that particular placement and operation can be arbitrary. For example, placing $\sigma(i, j)$ in between or before the transformed query and key matrices would also be valid as a re-weighting function application. $\sigma(i, j)$ can also map to any number of possible concentration methods, such as a matrix modifying $S(Q_{h,i}, K^T_{h,j})$ by multiplication or element-wise operations (e.g. addition).

Elaborate position-based encoding schemes \cite{raffel2020exploring, wang2020encoding, wang2020position, liutkus2021relative, alibi}, using absolute or relative token positions, have advanced the scheme utilized by 
the initial work~\cite{attention_is_all_you_need}
and many provide what can be intuited as position-based re-weighting functions. However, those schemes are specifically designed for a $S(Q_{h,i}, K^T_{h,j})$ formulation and do not work for the decomposed $S_q(Q_{h,i})S_k(K_{h,j}^T)$ linearized formulation. 

Rotary Positional Embeddings (RoPE) \cite{su2022roformer}, with some minor modifications, is closest to being a true position-based re-weighting function for linear transformers by using relative token positions\footnote[3]{Some re-weighting functions, like Alibi \cite{alibi}, are not covered in detail because we consider them superseded by other options or they are not obviously usable for linear attention because they are not decomposable like RoPE.}. However, RoPE is unaware of the total sequence length when it is applied, 
and this can cause potential 
problems. For example, RoPE would treat two tokens that are 100 tokens apart in a 1k length sequence and a 200 length sequence the same, where the actual relationship of the two tokens could vary drastically between the two sequences. This lack of sequence length awareness renders RoPE's re-weighting ability inherently limited, especially for sequences that exhibit more than the aforementioned locality characteristic. We elaborate on RoPE's construction (and its linear attention variant tested in this paper) in Appendix \ref{sec:rope}.

Recently introduced, cosFormer \cite{zhen2022cosformer} is the state-of-the-art in position-based linear transformers that utilizes sequence length in addition to absolute token positions. cosFormer's proposed mechanism, with common-sense modifications \cite{agostinelli2023improving}, is described by Equation \ref{eq:cos_gen_attn}. Here, $S_q$ and $S_k$ are set to $ReLU$ and their cosine-based re-weighting function is distributable via Ptolemy's method for expanding trigonometric expressions. Intuitively, when the positions $i$ and $j$ of two tokens are closer, the cosine's response is increased, emphasizing locality. Conversely, when the two positions are far apart, the response approaches zero, representing maximum attenuation via re-weighting. 



\vspace{-2em}

\begin{equation}
\vspace{-0.5em}
S(Q_{h,i}, K_{h,j}^T) = S_q(Q_{h,i})S_k(K_{h,j}^T)cos(\frac{\pi}{2}(\frac{i}{N_1} - \frac{j}{N_2}))
\label{eq:cos_gen_attn}
\end{equation}

Unlike RoPE, cosFormer can recognize differences in token distances relative to the sequence length, re-weighting more dynamically in practice. Using our previous example, cosFormer would treat two tokens that are 100 positions apart differently in a 1k length sequence versus a 200 length sequence, an intuitive improvement. 

\subsection{Motivation of Our Study}
Unfortunately, the reliance on sequence length 
makes it difficult to apply certain re-weighting functions towards autoregressive and simultaneous tasks. For instance, it can be challenging to apply position-based re-weighting functions to autoregressive tasks (e.g. text-to-speech translation) where target sequence lengths are usually not known beforehand. Although some effort has been made to address these issues \cite{liu2022neural, agostinelli2023improving}, mostly via target sequence length prediction based on the full input sequence, proposed solutions are prone to some level of approximation error. Furthermore, none of the prior approaches has 
discussed the impossibility of 
applying them to simultaneous tasks, where even the full input sequence is not available at decoding time-steps. 

Moreover, the static nature of the state-of-the-art re-weighting 
functions can cause issues from an inference quality standpoint. cosFormer's re-weighting function focuses on encouraging locality, but this can be problematic when locality bias is not important to a given application. RoPE and similar schemes suffer from the same problem. In such instances, dynamic flexibility in the re-weighting function to encourage strong, long-range connections would be preferred. An example of when this flexibility may be desirable can be found in a typical translation task, where languages like German that tend to exhibit subject-object-verb (SOV) structures as opposed to subject-verb-object (SVO) structures in languages like English may require diverse attention patterns and long-range dependencies. A verb near the end of a German sentence may attend strongly to the subject near the beginning of the sentence, but static re-weighting functions like the one employed by cosFormer would likely have trouble enabling this relationship.

\section{LeaPformer}
We propose a novel 
re-weighting function and method for constructing such functions for linear transformers that resolves 
the issues 
in applying them to many autoregressive tasks and enables their application to simultaneous tasks. 
To this end, 
we first generalize the reliance on absolute token position and sequence length into a more direct, intuitive reliance on the relative placement of a token in the sequence which we refer to as a \textit{proportion}. This generalization allows for easier analysis of re-weighting function behavior and removes theoretical dependence on sequence length. 
Second, we propose, construct, and deploy a compact module to learn proportional representations derived from each token, a technique that we call \textit{Learned Proportions} (LeaP) and call the models it is applied to LeaPformers. LeaPformers can be applied to tasks where sequence lengths are unknown and, more importantly, 
capture dynamic attention patterns over 
position-based re-weighting functions. 

\subsection{\emph{Proportion}-Based Re-weighting Functions}
We 
introduce proportion-based re-weighting in Equation \ref{eq:general_learned_cos_attn}, where $P_q$ and $P_k$ represent proportions of sequences from which queries and keys are derived from and $\sigma(P_{q, i}, P_{k, j})$ represents the re-weighting function with a reliance on the provided proportions. Technically, $P_q$ and $P_k$ can be set in any manner, but for the most straightforward proportion-based re-weighting implementations, they would correspond to the proportion of a sequence that a token is placed (e.g., at 20\% of the sequence). 

\vspace{-1.5em}

\begin{equation}
\begin{aligned}
&\quad \: P_q = [P_{q, 1}, \; P_{q, 2}, \: \dots \: \, P_{q, N_1}], \quad 0 \le \, P_{q, i} \le 1  \\
&\quad \: P_k = [P_{k, 1}, \; P_{k, 2}, \: \dots \: P_{k, N_2}], \quad 0 \le P_{k, j} \le 1  \\
&\quad S(Q_{h,i}, K_{h,j}^T) = S_q(Q_{h,i})S_k(K_{h,j}^T)\sigma(P_{q, i}, P_{k, j})
\end{aligned}
\label{eq:general_learned_cos_attn}
\end{equation}

Under this definition, cosFormer's formulation in Equation \ref{eq:cos_gen_attn} can be considered as a special case,
where we replace $\sigma(P_{q, i}, P_{k, j})$ in Equation \ref{eq:general_learned_cos_attn} with the cosine-based re-weighting function of cosFormer and define $P_q$ and $P_k$ as being explicit token positions divided by the sequence length, as shown in Equation \ref{eq:fixed_cos_attn}:

\vspace{-1.5em}

\begin{equation}
\vspace{-1em}
\begin{aligned}
&P_q = [\frac{1}{N_1}, \: \; \dots \:, \; \frac{N_1}{N_1}], \quad P_k = [\frac{1}{N_2}, \: \; \dots \: , \; \frac{N_2}{N_2}] \\
S(&Q_{h,i}, K_{h,j}^T) = S_q(Q_{h,i})S_k(K_{h,j}^T)cos(\frac{\pi}{2}(P_{q, i} - P_{k, j})) \\
\end{aligned}
\label{eq:fixed_cos_attn}
\end{equation}




\subsection{Learned Proportions}
In contrast to determining the proportions statically as in the case of cosFormer in Equation \ref{eq:fixed_cos_attn}, 
models can \emph{learn} to derive these representative proportions via a module containing a compact network embedded within attention blocks. We call this method \textit{Learned Proportions} (LeaP) and models utilizing this technique LeaPformers. The possible inference quality benefits of LeaP can be understood intuitively. Suppose that $P_k$ is set in a static manner in accordance with explicit positional representations, but $P_q$ is derived via a small module based on the query matrix. The module's learned behavior could produce derived elements of $P_q$ equal to classical positional representations, thus replicating the behavior and performance of attention mechanisms like cosFormer, but could alternatively defer the inter-token relationships that cosFormer might otherwise emphasize (i.e. an emphasis on locality). Along these lines, we 
redefine the aforementioned proportions in accordance with Equation \ref{eq:proportion_derivation}, where $LeaP_Q$ and $LeaP_K$ represent the proposed modules that derive proportions based on the query and key matrices, and $P_q$ and $P_k$ are redefined as $P_q(Q_h)$ and $P_k(K_h)$.


\begin{equation}
\vspace{-1em}
\begin{aligned}
&\qquad \qquad \; P_q(Q_{h,i}) \, = P_{q, i} \, = LeaP_Q(Q_{h,i}) \\
&\qquad \qquad \; P_k(K_{h,j}) = P_{k, j} = LeaP_K(K_{h,j}) \\
&P_q(Q_h) \, = [LeaP_Q(Q_{h,1}), \, \; \dots \;, \, LeaP_Q(Q_{h,N_1})] \\
&P_k(K_h) = [LeaP_K(K_{h,1}), \; \dots \;, LeaP_K(K_{h,N_2})] \\
\label{eq:proportion_derivation}
\end{aligned}
\vspace{-1em}
\end{equation}

To 
elaborate on potential inference quality benefits further, we can refer back to our example of translation to or from German and the SOV structure that cosFormer would likely struggle to model well. If $P_q$ is derived from a small LeaP module in self-attention, models could effectively defer the locality bias inherent to cosFormer to elsewhere in the sequence. If correctly learned, this might allow models to defer their attention concentration from the verb at the end of the German sequence to the beginning of the sequence, where we might expect a typically strong attention score. Allowing derivations of both $P_q$ and $P_k$ would, naturally, afford maximum flexibility in attention patterns produced by the employed re-weighting function. 


Beyond 
the inference quality benefits of LeaP, 
our method removes any dependence that proportion-based re-weighting functions have on knowing the sequence length beforehand, widely enabling them for autoregressive tasks without target sequence length prediction and, for the first time, 
demonstrating the feasibility to apply them to simultaneous tasks. 

\begin{figure}[t]
    \centering
    \includegraphics[scale=0.4]{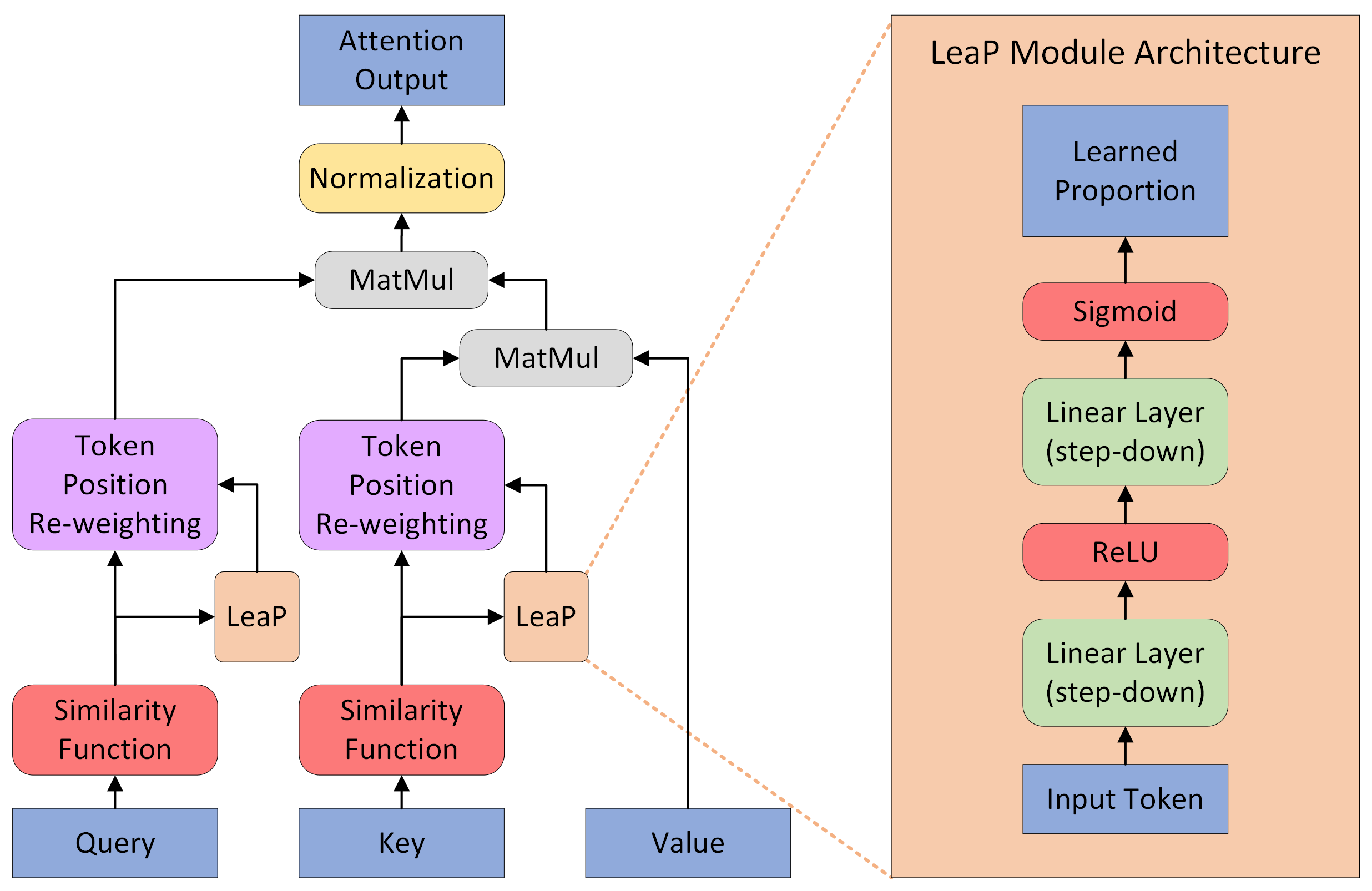}
    \vspace{-1em}
    \caption{Illustration of the proposed \textit{Learned Proportions} (LeaP) augmentation to linear transformer attention mechanisms. The LeaP module takes each token of the query and key matrices and 
    reduces their 
    dimensions 
    to a single proportion.}
    \label{fig:LeaP_arch}
\end{figure}

\subsection{Optimizing LeaP Module for Throughput and Analyzing Expressivity}
It is critical that the addition of LeaP does not significantly affect the throughput of a given model or its memory footprint, as it is intended for resource-constrained and latency-sensitive environments. Given that, we recommend a module composed of a simple, two-layer feed-forward network that steps down the attention head embedding dimension with a ReLU activation between the layers and a sigmoid activation at the end of the network, along the lines of the augmentation highlighted in Figure \ref{fig:LeaP_arch}. The choice of a ReLU activation is based on empirical tests on the Long-Range Arena benchmark \cite{tay2021long} which determined that, as opposed to several other competitive options, ReLU generalized well to multiple tasks. 

While a separate LeaP module for each attention head would be 
straightforward, we found in our 
experiments that this made a very minor difference in terms of quality. For English to German SimulST, we observed that when replacing the decoder self-attention block with LeaPformer where a separate LeaP module was provided for each attention head the models were of similar quality (
measured by validation perplexity, difference of 
$\sim$0.03). Given that and acknowledging that deploying multiple LeaP modules would drastically increase the parameter footprint of the module, we elect to share one LeaP module for all attention heads.

\begin{figure}[t]
    \centering
    \includegraphics[scale=0.31]{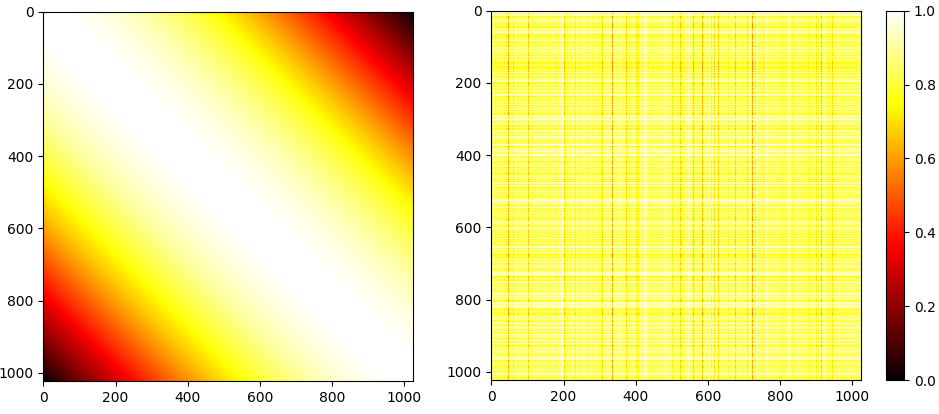}
    \vspace{-0.5em}
    \caption{An example of re-weighting matrices across all query (y-axis) and key (x-axis) token positions for baseline cosFormer (left) and LeaPformer (right) on list-operations in the Long-Range Arena benchmark. In this example, LeaPformer has clearly learned to attenuate more dynamically as opposed to the locality-focused, diagonalized re-weighting matrix of cosFormer.}
    \label{fig:LeaP_reweight}
\end{figure}

Additionally, given the activation functions chosen for the LeaP module's architecture, it is important to examine the expressivity of the module. It is generally desirable that the LeaP module outputs a complex range of values as opposed to saturating to values of 0 or 1, as otherwise it is simply sparsifying the $Q_hK_h^T$ matrix (were it to be directly calculated). 
In Figure \ref{fig:LeaP_reweight}, 
we compare the re-weighting matrices of baseline cosFormer and LeaPformer for the list-operations task in the LRA benchmark, a fairly difficult one. As can be observed in the example, and as we generally found in practice, the baseline cosFormer can only provide static re-weighting emphasizing locality (with the largest weights along diagonal for the same position). In contrast, cosFormer augmented with the LeaP modules is capable of generating complex re-weighting matrices that lightly attenuate between most positions while selectively attenuating harshly or not at all. The fact that there is wide-spread, light attenuation across several examples indicates that our LeaP module 
can also avoid saturation. 



\begin{table*}[t]
    \vspace{-.8em}      
    \centering
    \caption{
    Results on the Long-Range Arena benchmark. 
    We measure the accuracy (higher is better) and are weighted evenly for the purpose of the average score. Best results are \textbf{bolded}; 
    runner-up results are \underline{underlined}. 
    LeaPformers 
    showcase competitive performance across a range of tasks. 
    LeaPformer-1.5\% achieves the second best average score, beating all other non-quadratic transformers.}
    \vspace{0.5em}
    \begin{tabular}{l|ccccc|c}
         \toprule
         Attention Mechanism & ListOps & Text Cls. & Text Rtr. & Path-32 & Img. Cls. & Avg. \\
         \midrule
         Softmax Attn. \cite{attention_is_all_you_need} &  37.94 & 60.51 & 80.52 & \textbf{75.54} & \underline{41.74} & \textbf{59.25} \\
         Linear Attn. \cite{kathrapalous20} & \textbf{39.21} & 61.53 & 78.78 & 68.23 & 39.14 & 57.38 \\
         Linformer \cite{linformer2020} & 37.04 & 57.65 & 77.61 & 57.91 & 37.85 & 53.61 \\
         Performer \cite{performer2020} & 38.17 & 64.24 & 80.11 & 68.54 & 37.42 & 57.70 \\
         BigBird \cite{bigbird2020} &  38.36 & 60.72 & \underline{80.97} & \underline{72.80} & 40.37 & 58.64 \\
         Reformer \cite{reformer2020} & 36.44 & 63.14 & 78.63 & 69.29 & \textbf{42.85} & 58.07 \\
         Skyformer \cite{chen2021skyformer} & 38.66 & \textbf{65.38} & \textbf{81.77} & 68.74 & 36.07 & 58.12 \\
         RoPE w/ Linear Attn. \cite{su2022roformer} & 38.31 & 64.79 & 77.54 & 67.61 & 39.17 & 57.48 \\
         cosFormer \cite{zhen2022cosformer} & \underline{38.96} & 61.66 & 79.29 & 68.96 & 38.26 & 57.43 \\
         \midrule
         LeaPformer-0.2\% & 38.26 & 64.70 & 79.88 & 70.76 & 38.26 & 58.37 \\
         LeaPformer-1.5\% & \underline{38.96} & \underline{64.90} & 80.62 & 68.99 & 40.00 & \underline{58.69} \\
         \bottomrule
    \end{tabular}
    \label{tab:lra_results_quality}
\end{table*}

\section{Empirical Evaluation}
We validate the potential of LeaP by applying it to cosFormer on three major sets of tasks. 
All references to LeaPformers in the following sections refer to this augmentation of cosFormers. We first test LeaPformers on the popular Long-Range Arena (LRA) benchmark \cite{tay2021long}, built specifically for validating the capabilities of efficient attention mechanisms. We also engage with basic autoregressive language modeling, employing \citealt{baevski2019adaptive}'s adaptive input/output architecture on Wikitext-103b \cite{merity2016pointer}. 
Moreover, we evaluate LeaPformers on speech-to-text simultaneous translation (SimulST) via a wait-k read-write schedule \cite{mma19, mono_xma20, simulst_xma20} across two language pairs. For our SimulST and autoregressive language modeling experiments, we employ Fairseq \cite{fairseq2019} for training and validation alongside SimulEval \cite{simuleval2020} for SimulST evaluation. LRA results are compared via accuracy, autoregressive language modeling results are evaluated via validation and test set perplexity, and SimulST results are compared via detokenized BLEU-4 (called BLEU later) using sacreBLEU \cite{sacrebleu2018}. Additional details related to employed hardware and hyperparameters can be found in the Appendix.

\subsection{Long-Range Arena Benchmark}
Instead of the Long-Range Arena (LRA) benchmark provided by \citealt{tay2021long}, our implementation follows Skyformer's \cite{chen2021skyformer} PyTorch framework and reuses their architectures and hyperparameters, which we hold static. We provide baseline results for 
various architectures, including the classical transformer \cite{attention_is_all_you_need} and several seminal efficient transformers. Some auxiliary results and details related to the LRA benchmark are provided in Appendices \ref{sec:repr_cap} and \ref{sec:eff_design} regarding controlling for increased parameter counts and alternatives to efficient transformers.

In addition to these results, we propose a composite heuristic evaluation metric that we call \textit{Relative Composite Performance (RCP)} to more concretely evaluate efficient attention mechanisms and their throughput-accuracy trade-offs. We treat softmax attention as an inference quality ceiling and throughput floor for the LRA benchmark 
as follows:
\begin{equation}
    RCP(ef, sft, std_{bench}) = \frac{ef_{thrpt} / sft_{thrpt}}{(1 + \frac{sft_{acc} - eft_{acc}}{std_{bench}})}
    \label{eq:rcp}
\end{equation}
with $ef$ as a given efficient attention mechanism and $sft$ as softmax attention and their corresponding accuracy and throughput values. The RCP numerator rewards significant speedups in a proportional manner via a simple ratio. Contrastingly, the RCP denominator is governed by the delta in accuracy between two attention mechanisms normalized by the standard deviation of the entire benchmark's accuracy, focusing on penalizing for inaccuracy. Adding by one in the denominator smooths out the resulting values. While $RPC$ as depicted in Equation \ref{eq:rcp} favors equally prioritizing accuracy and throughput, one could prioritize one or the other by changing the exponential values of the expressions in the numerator and denominator (they are currently set to 1 as a default for equal prioritization). 


Equation \ref{eq:rcp_mem} 
is a memory footprint-aware version of the metric, called RCP$_{mem}$ that splits its reward between throughput increases and memory footprint reductions, where the weights for those rewards are similarly tunable:
%
%
\begin{equation}
    RCP_{mem}(ef, sft, std_{bench}) = \frac{\frac{1}{2}(\frac{ef_{thrpt}}{sft_{thrpt}}) + \frac{1}{2}(\frac{sft_{mem}}{ef_{mem}})}{(1 + \frac{sft_{acc} - eft_{acc}}{std_{bench}})}
    \label{eq:rcp_mem}
\end{equation}
We set these tunable weights to 0.5 
as a default.

Regarding the LeaPformers tested on the LRA benchmark, a minimal setup was initially employed with around a maximum of a 0.2\% increase on the number of parameters for the LeaP module. 
We additionally test a larger module 
employed with a maximum increase of 1.5\% to the number of parameters to investigate the effects of increased size. Some very limited fine-tuning was employed across a few possible module sizes on a per-task basis for the larger LeaPformer, depending on the perceived difficulty of the task.

We show 
a holistic view of performance 
in Figure \ref{fig:trade-off} (we refer the readers to \S\ref{sec:intro}), with kernel-based linear transformers tending to provide an excellent quality-throughput trade-off. 
It is clear from the figure that 
LeaPformer provides the best performance trade-off, exhibiting significant quality increases over Linear Transformer and overall supremacy compared to Performer, Linformer, Reformer, and Skyformer, with a 
reduced memory footprint. Details on inference quality are showcased in Table \ref{tab:lra_results_quality}, where both LeaPformer-0.2\% and LeaPformer-1.5\% exhibit a balanced performance profile. While classical softmax attention achieves the highest average score by a notable margin, it is beaten on a number of tasks by other methods. 



\begin{table}[t]
    \vspace{-.8em}  
    \centering
    \caption{
    Throughput comparison on the Long-Range Arena benchmark. Training throughput values (higher is better, inference speed is identical) are provided for sequence lengths in \{1k, 2k, 4k\}. 
    The results are ordered from fastest to slowest according to their 4k sequence length throughput. LeaPformers are \textbf{bolded}.}
    \vspace{0.5em}
    \begin{tabular}{l|ccc}
        \toprule
        Attention Mechanism & \multicolumn{3}{c}{Training Thrpt. (itr/sec)} \\
         & 1k & 2k & 4k \\
        \midrule
         Linear Attn. & 68.00 & 28.43 & 15.18 \\
         cosFormer & 58.91 & 25.64 & 13.13 \\
         \textbf{LeaPformer-0.2\%} & 56.30 & 24.72 & 12.81 \\
         RoPE w/ Linear Attn. & 48.90 & 23.81 & 12.27\\
         \textbf{LeaPformer-1.5\%} & 53.58 & 23.39 & 11.76 \\
         Linformer & 48.96 & 20.49 & 11.36 \\
         Performer & 38.83 & 17.72 & \hphantom{0}9.02 \\
         Reformer & 32.07 & 15.08 & \hphantom{0}7.71 \\
         Skyformer & 26.02 & 12.36 & \hphantom{0}6.06 \\
         BigBird & 15.97 & \hphantom{0}6.76 & \hphantom{0}3.52 \\
         Softmax Attn. & 14.08 & \hphantom{0}6.03 & \hphantom{0}1.64 \\
         \bottomrule
    \end{tabular}
    \label{tab:lra_results_latency}
    \vspace{-1.25em}
\end{table}

Compared to cosFormer, LeaPformer provides, at a minor throughput and memory footprint penalty, significant increases to scores across cosFormer's most problematic tasks, including improvements for text and image classification. Additionally, when compared to the score profiles of other efficient attention mechanisms, LeaPformer does not seem to specialize nearly as much as other architectures (aside from some difficulty on the pathfinding task), indicating its balanced performance. BigBird is the closest to providing a similarly balanced inference quality profile, but this comes with significant throughput reductions as shown in Table \ref{tab:lra_results_latency} and noticeable increases to memory footprint. Regarding the application of RCP to the results in Table \ref{tab:lra_results_quality} and Table \ref{tab:lra_results_latency}, LeaPformer beats out all other options in Table \ref{tab:rcp_results} by a wide margin (minimum increase of 0.81 RCP), demonstrating its very effective relative performance on the LRA benchmark compared to softmax attention and its efficient attention peers. Across the board, LeaPformer matches the general inference quality of task-balanced models with a massively reduced memory footprint while still exhibiting a minimum 1.52x throughput increase over those mechanisms.


\subsection{Autoregressive Language Modeling}

\begin{table}[t]
    \vspace{-.8em}
    \centering
    \caption{Results for LRA 
    using the proposed metric \textit{Relative Composite Performance} (RCP) treating softmax attention as a throughput floor and accuracy ceiling. Throughput values for RCP calculation are derived from 4k sequence tasks and the accuracy and standard deviation of accuracy are derived from the average accuracy across all five tasks in the LRA benchmark. RCP scores are sorted from highest to lowest with RCP$_{mem}$ provided as an auxiliary option that rewards reduced memory footprints. LeaPformers are \textbf{bolded}. Softmax attention is naturally excluded, as the following RCP values are relative to it.}
    \vspace{0.5em}
    \begin{tabular}{l|cc}
         \toprule
         Attention Mechanism & RCP & RCP$_{mem}$ \\
         \midrule
         \textbf{LeaPformer-1.5\%} & 5.20 & 5.04 \\
         \textbf{LeaPformer-0.2\%} & 4.90 & 4.83 \\
         Linear Attn. & 4.09 & 4.33 \\
         cosFormer & 3.59 & 3.54 \\
         RoPE w/ Linear Attn. & 3.41 & 3.68 \\
         Performer & 2.69 & 2.57 \\
         Reformer & 2.62 & 2.31 \\
         Skyformer &  2.10 & 2.02 \\
         BigBird & 1.52 & 1.53 \\
         Linformer & 1.44 & 1.31 \\
         \bottomrule
    \end{tabular}
    \label{tab:rcp_results}
    \vspace{-.8em}
\end{table}

While autoregressive language modeling has advanced tremendously with the advent of LLMs, more accessible methods can still serve to validate architectural differences between attention mechanisms. Given that, we've employed the adaptive setup of \citealt{baevski2019adaptive} and reuse nearly their exact model hyperparameters for autoregressive language modeling on Wikitext-103b \cite{merity2016pointer}. Hyperparameter differences are only related to batch sizes and the number of updates due to computational constraints, and are detailed in our Appendix. All sequences during training and evaluation were composed of 512 tokens (i.e. 511 tokens of context where possible for evaluation). 

\begin{table}[h]
    \vspace{-.8em}
    \centering
    \caption{Results from autoregressive language modeling on Wikitext-103b. 
    Softmax attention is a baseline. 
    Perplexity (ppl; lower is better) is measured on the recommended dev and test sets.}
    \vspace{0.5em}
    \begin{tabular}{l|c|c}
        \toprule
         Attention Mechanism & ppl(dev) & ppl(test) \\
         \midrule
         Softmax Attention & 22.53 & 21.67 \\
         \midrule
         ReLU Attention & 24.55 & 24.16 \\
         cosFormer & 24.50 & 24.17 \\
         LeaPformer & 24.46 & 24.04 \\ \bottomrule
    \end{tabular}
    \label{tab:auto_lmm}
    \vspace{-.8em}
\end{table}

We contrast a few attention mechanisms in Table \ref{tab:auto_lmm}, including a ReLU-
based mechanism functioning similar to Linear Transformers \cite{kathrapalous20} as well as an ablation of cosFormer's re-weighting function \cite{zhen2022cosformer}. As observed in Table \ref{tab:auto_lmm}, classical softmax attention outperforms all linear attention mechanisms by a wide margin, but amongst the linear attention mechanisms themselves there are distinctions in terms of quality. Notably, LeaPformer demonstrates significant improvement over its linear attention peers, beating out cosFormer by 0.13 perplexity on the test set while only requiring a parameter increase of approximately 3.13\%, though it still falls significantly short of classical softmax attention by 2.37 perplexity. 

\begin{table}[t]
    \vspace{-.8em}
    \centering
    \caption{
    Comparison of cosFormer and LeaPformer trained on MuST-C en-de for SimulST. Perplexity (ppl; lower is better) is measured on the training and validation sets. BLEU scores are not provided for baseline cosFormer as it is impossible to apply to 
    the tasks (i.e. the scores become zero) without augmentations.}
    \vspace{0.5em}
    \begin{tabular}{l|c|c}
        \toprule
        Attention Mechanism & ppl(tr) & ppl(dev) \\
        \midrule
        cosFormer Dec. Self-Attn. & 8.44 & 9.86 \\
        LeaPformer Dec. Self-Attn. & 7.86 & 9.40 \\ \bottomrule
    \end{tabular}
    \label{tab:LeaP_vs_cos}
\end{table}

\subsection{Simultaneous Speech Translation (SimulST)}


For the purposes of our SimulST related experiments, we employed a model inspired by the ESPnet-ST toolkit \cite{espnetst2020} that focused on end-to-end speech-to-text (S2T) translation with a modified cross-attention block for a wait-k and fixed pre-decision paradigm \cite{mma19,mono_xma20,simulst_xma20}. All model encoders were pre-trained on automatic speech-recognition (ASR) and were trained on a wait-k of 5 and a fixed predecision ratio of 9 and were evaluated on a wait-k of 3 (a slightly larger k for training is suggested by prior work \citep{mma19}) with greedy decoding. Models are evaluated via validation set perplexity and by detokenized BLEU-4 \cite{sacrebleu2018} via SimulEval \cite{simuleval2020}. Two language pairs and two datasets were employed to test the application of LeaPformer to simultaneous tasks. We utilized MuST-C's \cite{mustc2021} English to German (en-de) split and CoVoST 2's \cite{wang2020covost} French to English (fr-en) split. More comprehensive evaluation is provided for the en-de pair, comparing the results of LeaPformer to an ablation without a re-weighting function. The application of LeaP modules resulted in an approximate parameter increase of 0.03\% for ablations that included all attention blocks being linearized.



\begin{table}[h]
    \vspace{-.8em}
    \centering
    \caption{Results from SimulST for MuST-C en-de for various LeaPformer and simple ReLU ablations with softmax as a baseline. BLEU scores (higher is better) are generated on on tst-COMMON.}
    \vspace{0.5em}
    \begin{tabular}{l|c|c}
        \toprule
         Attention Mechanism & BLEU & ppl(dev) \\
         \midrule
         Softmax Attention & 15.07 & \hphantom{0}9.36 \\
         \midrule
         LeaPformer Enc. Self-Attn. & 12.00 & 11.50 \\
         LeaPformer Dec. Self-Attn. & 14.81 & \hphantom{0}9.40 \\
         LeaPformer Cross-Attn. & 13.95 & 11.02 \\
         LeaPformer All Attn. & 11.19 & 14.67 \\
         \midrule
         ReLU Enc. Self-Attn. & 11.55 & 11.98 \\
         ReLU Dec. Self-Attn. & 14.67 & \hphantom{0}9.55 \\
         ReLU Dec. Cross-Attn. & 13.84 & 11.24 \\
         ReLU All Attn. & 10.38 & 15.48 \\ \bottomrule
    \end{tabular}
    \label{tab:simulst_ablation}
\end{table}

We first seek to show LeaPformer outperforms cosFormer in terms of model quality, justifying its inclusion not only from the perspective of necessity but also as an overall improvement. Table \ref{tab:LeaP_vs_cos} demonstrates the results of a brief comparison on en-de SimulST (note that cosFormer can still be employed for training, where  sequence lengths are known), where significant quality improvement is observed. Having established the capability of the proposed method, we seek to validate it further on en-de SimulST while also providing several ablations for LeaPformer, representing a wide-range of quality-throughput trade-offs. Additionally, we seek to show that applying the LeaP-augmented re-weighting function of LeaPformer is consistently useful by testing models trained without any re-weighting functionality, operating as a variation on Linear Transformer \cite{kathrapalous20}. Table \ref{tab:simulst_ablation} showcases the results of this study, where LeaPformer ablations consistently beat their simple ReLU-based alternative. The most competitive ablation in terms of translation quality emerges as a model with the decoder self-attention block replaced by LeaPformer, achieving only a 0.26 BLEU reduction compared to softmax attention.

\begin{table}[t]
    \vspace{-.8em}
    \centering
    \caption{Results from SimulST for CoVoST fr-en for various LeaPformer ablations with softmax as a baseline. BLEU scores are generated on the recommended, but shortened, test split.}
    \vspace{0.5em}
    \begin{tabular}{l|c|c}
        \toprule
         Attention Mechanism & BLEU & ppl(dev) \\
         \midrule
         Softmax Attention & 14.51 & 9.99 \\
         \midrule
         LeaPformer Enc. Self-Attn. & 11.18 & 12.50 \\
         LeaPformer Dec. Self-Attn. & 14.28 & 10.11 \\
         LeaPformer Cross-Attn. & 13.25 & 11.64 \\
         LeaPformer All Attn. & \hphantom{0}9.69 & 16.28 \\ \bottomrule
    \end{tabular}
    \label{tab:simulst_fr_en}
\end{table}

Similar results are provided for the fr-en language pair in Table \ref{tab:simulst_fr_en}, with trends from en-de persisting. The most competitive translation quality ablations continue to be replacements of the decoder self-attention blocks with LeaPformer, where only a 0.23 BLEU reduction was observed. While our analysis related to SimulST is focused on analyzing the possible translation quality benefits of employing LeaPformers, we also provide some latency analysis, with some qualifications, employing common SimulST latency metrics in Appendix \ref{sec:simulst_latency}. 


\section{Conclusion}
In this paper, we made two concrete contributions. We re-framed reliance on explicit positional representations and sequence lengths to reliance on sequence proportions, removing theoretical dependence on sequence lengths. Additionally, we proposed LeaPformers and applied them to the state-of-the-art in proportion-based linear transformers, cosFormer, achieving the best performance trade-off on the Long-Range Arena benchmark and competitive results in autoregressive language modeling on Wikitext-103b. Moreover, we applied proportion-based transformers for the first time to simultaneous translation, achieving minimal quality loss compared to softmax attention for two language pairs.

\section{Impact Statement}
We advance the efficiency of transformers in state-of-the-art deep learning. Any societal consequences or impacts that typically relate to work focused on increased efficiency also apply here, as such work necessarily improves the practicality of deep learning models for an array of applications.


\section*{Acknowledgement}
We thank the anonymous reviewers for valuable feedback.
This research was supported, in part, by the National Science Foundation grants 2223483 and 2223484.
Sanghyun is also partially supported by the Google Faculty Research Award.

\bibliography{custom}

\begin{thebibliography}{43}
\providecommand{\natexlab}[1]{#1}
\providecommand{\url}[1]{\texttt{#1}}
\expandafter\ifx\csname urlstyle\endcsname\relax
  \providecommand{\doi}[1]{doi: #1}\else
  \providecommand{\doi}{doi: \begingroup \urlstyle{rm}\Url}\fi

\bibitem[Agostinelli \& Chen(2023)Agostinelli and Chen]{agostinelli2023improving}
Agostinelli, V. and Chen, L.
\newblock Improving autoregressive nlp tasks via modular linearized attention.
\newblock In Koutra, D., Plant, C., Gomez~Rodriguez, M., Baralis, E., and Bonchi, F. (eds.), \emph{Machine Learning and Knowledge Discovery in Databases: Research Track}, pp.\  90--106, Cham, 2023. Springer Nature Switzerland.
\newblock ISBN 978-3-031-43421-1.

\bibitem[Ainslie et~al.(2020)Ainslie, Ontanon, Alberti, Cvicek, Fisher, Pham, Ravula, Sanghai, Wang, and Yang]{etc_ainslie2020}
Ainslie, J., Ontanon, S., Alberti, C., Cvicek, V., Fisher, Z., Pham, P., Ravula, A., Sanghai, S., Wang, Q., and Yang, L.
\newblock Etc: Encoding long and structured inputs in transformers, 2020.
\newblock URL \url{https://arxiv.org/abs/2004.08483}.

\bibitem[Baevski \& Auli(2019)Baevski and Auli]{baevski2019adaptive}
Baevski, A. and Auli, M.
\newblock Adaptive input representations for neural language modeling.
\newblock In \emph{International Conference on Learning Representations}, 2019.
\newblock URL \url{https://openreview.net/forum?id=ByxZX20qFQ}.

\bibitem[Beltagy et~al.(2020)Beltagy, Peters, and Cohan]{longformer2020}
Beltagy, I., Peters, M.~E., and Cohan, A.
\newblock Longformer: The long-document transformer, 2020.
\newblock URL \url{https://arxiv.org/abs/2004.05150}.

\bibitem[Cattoni et~al.(2021)Cattoni, {Di Gangi}, Bentivogli, Negri, and Turchi]{mustc2021}
Cattoni, R., {Di Gangi}, M.~A., Bentivogli, L., Negri, M., and Turchi, M.
\newblock Must-c: A multilingual corpus for end-to-end speech translation.
\newblock \emph{Computer Speech \& Language}, 66:\penalty0 101155, 2021.
\newblock ISSN 0885-2308.
\newblock \doi{https://doi.org/10.1016/j.csl.2020.101155}.
\newblock URL \url{https://www.sciencedirect.com/science/article/pii/S0885230820300887}.

\bibitem[Chen et~al.(2021)Chen, Zeng, Ji, and Yang]{chen2021skyformer}
Chen, Y., Zeng, Q., Ji, H., and Yang, Y.
\newblock Skyformer: Remodel self-attention with gaussian kernel and nystr\"om method.
\newblock In \emph{Advances in Neural Information Processing Systems 35: Annual Conference on Neural Information Processing Systems 2021, NeurIPS 2021, December 6-14, 2021, virtual}, 2021.

\bibitem[Child et~al.(2019)Child, Gray, Radford, and Sutskever]{sparsetransformer}
Child, R., Gray, S., Radford, A., and Sutskever, I.
\newblock Generating long sequences with sparse transformers, 2019.
\newblock URL \url{https://arxiv.org/abs/1904.10509}.

\bibitem[Choromanski et~al.(2020)Choromanski, Likhosherstov, Dohan, Song, Gane, Sarlos, Hawkins, Davis, Mohiuddin, Kaiser, Belanger, Colwell, and Weller]{performer2020}
Choromanski, K., Likhosherstov, V., Dohan, D., Song, X., Gane, A., Sarlos, T., Hawkins, P., Davis, J., Mohiuddin, A., Kaiser, L., Belanger, D., Colwell, L., and Weller, A.
\newblock Rethinking attention with performers, 2020.
\newblock URL \url{https://arxiv.org/abs/2009.14794}.

\bibitem[Dai et~al.(2019)Dai, Yang, Yang, Carbonell, Le, and Salakhutdinov]{transformerxl}
Dai, Z., Yang, Z., Yang, Y., Carbonell, J., Le, Q.~V., and Salakhutdinov, R.
\newblock Transformer-xl: Attentive language models beyond a fixed-length context, 2019.
\newblock URL \url{https://arxiv.org/abs/1901.02860}.

\bibitem[Dosovitskiy et~al.(2021)Dosovitskiy, Beyer, Kolesnikov, Weissenborn, Zhai, Unterthiner, Dehghani, Minderer, Heigold, Gelly, Uszkoreit, and Houlsby]{dosovitskiy2021image}
Dosovitskiy, A., Beyer, L., Kolesnikov, A., Weissenborn, D., Zhai, X., Unterthiner, T., Dehghani, M., Minderer, M., Heigold, G., Gelly, S., Uszkoreit, J., and Houlsby, N.
\newblock An image is worth 16x16 words: Transformers for image recognition at scale, 2021.

\bibitem[Gu \& Dao(2023)Gu and Dao]{gu2023mamba}
Gu, A. and Dao, T.
\newblock Mamba: Linear-time sequence modeling with selective state spaces, 2023.

\bibitem[Gu et~al.(2022)Gu, Goel, and R\'e]{gu2022efficiently}
Gu, A., Goel, K., and R\'e, C.
\newblock Efficiently modeling long sequences with structured state spaces.
\newblock In \emph{The International Conference on Learning Representations ({ICLR})}, 2022.

\bibitem[Inaguma et~al.(2020)Inaguma, Kiyono, Duh, Karita, Soplin, Hayashi, and Watanabe]{espnetst2020}
Inaguma, H., Kiyono, S., Duh, K., Karita, S., Soplin, N. E.~Y., Hayashi, T., and Watanabe, S.
\newblock Espnet-st: All-in-one speech translation toolkit, 2020.
\newblock URL \url{https://arxiv.org/abs/2004.10234}.

\bibitem[Katharopoulos et~al.(2020)Katharopoulos, Vyas, Pappas, and Fleuret]{kathrapalous20}
Katharopoulos, A., Vyas, A., Pappas, N., and Fleuret, F.
\newblock Transformers are rnns: Fast autoregressive transformers with linear attention, 2020.
\newblock URL \url{https://arxiv.org/abs/2006.16236}.

\bibitem[Kitaev et~al.(2020)Kitaev, Kaiser, and Levskaya]{reformer2020}
Kitaev, N., Kaiser, L., and Levskaya, A.
\newblock Reformer: The efficient transformer, 2020.
\newblock URL \url{https://arxiv.org/abs/2001.04451}.

\bibitem[Liu et~al.(2022)Liu, Li, Lu, Qin, Sun, Xu, and Zhong]{liu2022neural}
Liu, Z., Li, D., Lu, K., Qin, Z., Sun, W., Xu, J., and Zhong, Y.
\newblock Neural architecture search on efficient transformers and beyond, 2022.

\bibitem[Liutkus et~al.(2021)Liutkus, C{\'i}fka, Wu, {\c S}im{\c s}ekli, Yang, and Richard]{liutkus2021relative}
Liutkus, A., C{\'i}fka, O., Wu, S.-L., {\c S}im{\c s}ekli, U., Yang, Y.-H., and Richard, G.
\newblock Relative positional encoding for {Transformers} with linear complexity.
\newblock In Meila, M. and Zhang, T. (eds.), \emph{Proceedings of the 38th International Conference on Machine Learning}, volume 139 of \emph{Proceedings of Machine Learning Research}, pp.\  7067--7079. PMLR, 18--24 Jul 2021.
\newblock URL \url{http://proceedings.mlr.press/v139/liutkus21a.html}.

\bibitem[Ma et~al.(2019)Ma, Huang, Xiong, Zheng, Liu, Zheng, Zhang, He, Liu, Li, Wu, and Wang]{mma19}
Ma, M., Huang, L., Xiong, H., Zheng, R., Liu, K., Zheng, B., Zhang, C., He, Z., Liu, H., Li, X., Wu, H., and Wang, H.
\newblock Stacl: Simultaneous translation with implicit anticipation and controllable latency using prefix-to-prefix framework.
\newblock In \emph{Proceedings of the 57th Annual Meeting of the Association for Computational Linguistics}, pp.\  3025--3036, Florence, Italy, 2019. Association for Computational Linguistics (ACL).

\bibitem[Ma et~al.(2020{\natexlab{a}})Ma, Dousti, Wang, Gu, and Pino]{simuleval2020}
Ma, X., Dousti, M.~J., Wang, C., Gu, J., and Pino, J.
\newblock Simuleval: An evaluation toolkit for simultaneous translation, 2020{\natexlab{a}}.
\newblock URL \url{https://arxiv.org/abs/2007.16193}.

\bibitem[Ma et~al.(2020{\natexlab{b}})Ma, Pino, Cross, Puzon, and Gu]{mono_xma20}
Ma, X., Pino, J., Cross, J., Puzon, L., and Gu, J.
\newblock Monotonic multihead attention.
\newblock In \emph{International Conference on Learning Representations}, 2020{\natexlab{b}}.

\bibitem[Ma et~al.(2020{\natexlab{c}})Ma, Pino, Cross, Puzon, and Gu]{simulst_xma20}
Ma, X., Pino, J., Cross, J., Puzon, L., and Gu, J.
\newblock Simulmt to simulst: Adapting simultaneous text translation to end-to-end simultaneous speech translation.
\newblock In \emph{Proceedings of 2020 Asia-Pacific Chapter of the Association for Computational Linguistics and the International Joint Conference on Natural Language Processing}, 2020{\natexlab{c}}.

\bibitem[Merity et~al.(2016)Merity, Xiong, Bradbury, and Socher]{merity2016pointer}
Merity, S., Xiong, C., Bradbury, J., and Socher, R.
\newblock Pointer sentinel mixture models.
\newblock \emph{ArXiv}, abs/1609.07843, 2016.
\newblock URL \url{https://api.semanticscholar.org/CorpusID:16299141}.

\bibitem[Ott et~al.(2019)Ott, Edunov, Baevski, Fan, Gross, Ng, Grangier, and Auli]{fairseq2019}
Ott, M., Edunov, S., Baevski, A., Fan, A., Gross, S., Ng, N., Grangier, D., and Auli, M.
\newblock fairseq: A fast, extensible toolkit for sequence modeling, 2019.
\newblock URL \url{https://arxiv.org/abs/1904.01038}.

\bibitem[Parmar et~al.(2018)Parmar, Vaswani, Uszkoreit, Kaiser, Shazeer, Ku, and Tran]{image_transf}
Parmar, N., Vaswani, A., Uszkoreit, J., Kaiser, L., Shazeer, N., Ku, A., and Tran, D.
\newblock Image transformer, 2018.
\newblock URL \url{https://arxiv.org/abs/1802.05751}.

\bibitem[Peng et~al.(2021)Peng, Pappas, Yogatama, Schwartz, Smith, and Kong]{rfapeng2021}
Peng, H., Pappas, N., Yogatama, D., Schwartz, R., Smith, N.~A., and Kong, L.
\newblock Random feature attention, 2021.
\newblock URL \url{https://arxiv.org/abs/2103.02143}.

\bibitem[Post(2018)]{sacrebleu2018}
Post, M.
\newblock A call for clarity in reporting {BLEU} scores.
\newblock In \emph{Proceedings of the Third Conference on Machine Translation: Research Papers}, pp.\  186--191, Belgium, Brussels, October 2018. Association for Computational Linguistics.
\newblock URL \url{https://www.aclweb.org/anthology/W18-6319}.

\bibitem[Press et~al.(2022)Press, Smith, and Lewis]{alibi}
Press, O., Smith, N., and Lewis, M.
\newblock Train short, test long: Attention with linear biases enables input length extrapolation.
\newblock In \emph{International Conference on Learning Representations}, 2022.
\newblock URL \url{https://openreview.net/forum?id=R8sQPpGCv0}.

\bibitem[Qin et~al.(2022{\natexlab{a}})Qin, Han, Sun, Li, Kong, Barnes, and Zhong]{qin-etal-2022-devil}
Qin, Z., Han, X., Sun, W., Li, D., Kong, L., Barnes, N., and Zhong, Y.
\newblock The devil in linear transformer.
\newblock In \emph{Proceedings of the 2022 Conference on Empirical Methods in Natural Language Processing}, pp.\  7025--7041, Abu Dhabi, United Arab Emirates, December 2022{\natexlab{a}}. Association for Computational Linguistics.
\newblock URL \url{https://aclanthology.org/2022.emnlp-main.473}.

\bibitem[Qin et~al.(2022{\natexlab{b}})Qin, Sun, Deng, Li, Wei, Lv, Yan, Kong, and Zhong]{zhen2022cosformer}
Qin, Z., Sun, W., Deng, H., Li, D., Wei, Y., Lv, B., Yan, J., Kong, L., and Zhong, Y.
\newblock cosformer: Rethinking softmax in attention.
\newblock In \emph{International Conference on Learning Representations}, 2022{\natexlab{b}}.
\newblock URL \url{https://openreview.net/forum?id=Bl8CQrx2Up4}.

\bibitem[Raffel et~al.(2020)Raffel, Shazeer, Roberts, Lee, Narang, Matena, Zhou, Li, and Liu]{raffel2020exploring}
Raffel, C., Shazeer, N., Roberts, A., Lee, K., Narang, S., Matena, M., Zhou, Y., Li, W., and Liu, P.~J.
\newblock Exploring the limits of transfer learning with a unified text-to-text transformer.
\newblock \emph{J. Mach. Learn. Res.}, 21\penalty0 (1), jan 2020.
\newblock ISSN 1532-4435.

\bibitem[Su et~al.(2022)Su, Lu, Pan, Murtadha, Wen, and Liu]{su2022roformer}
Su, J., Lu, Y., Pan, S., Murtadha, A., Wen, B., and Liu, Y.
\newblock Roformer: Enhanced transformer with rotary position embedding, 2022.

\bibitem[Tay et~al.(2020)Tay, Bahri, Yang, Metzler, and Juan]{tay2020sparse}
Tay, Y., Bahri, D., Yang, L., Metzler, D., and Juan, D.-C.
\newblock Sparse sinkhorn attention, 2020.
\newblock URL \url{https://arxiv.org/abs/2002.11296}.

\bibitem[Tay et~al.(2021)Tay, Dehghani, Abnar, Shen, Bahri, Pham, Rao, Yang, Ruder, and Metzler]{tay2021long}
Tay, Y., Dehghani, M., Abnar, S., Shen, Y., Bahri, D., Pham, P., Rao, J., Yang, L., Ruder, S., and Metzler, D.
\newblock Long range arena : A benchmark for efficient transformers.
\newblock In \emph{International Conference on Learning Representations}, 2021.
\newblock URL \url{https://openreview.net/forum?id=qVyeW-grC2k}.

\bibitem[Tay et~al.(2022)Tay, Dehghani, Bahri, and Metzler]{tay2022efficient}
Tay, Y., Dehghani, M., Bahri, D., and Metzler, D.
\newblock Efficient transformers: A survey.
\newblock \emph{ACM Comput. Surv.}, 55\penalty0 (6), dec 2022.
\newblock ISSN 0360-0300.
\newblock \doi{10.1145/3530811}.
\newblock URL \url{https://doi.org/10.1145/3530811}.

\bibitem[Vaswani et~al.(2017)Vaswani, Shazeer, Parmar, Uszkoreit, Jones, Gomez, Kaiser, and Polosukhin]{attention_is_all_you_need}
Vaswani, A., Shazeer, N., Parmar, N., Uszkoreit, J., Jones, L., Gomez, A.~N., Kaiser, L.~u., and Polosukhin, I.
\newblock Attention is all you need.
\newblock In Guyon, I., Luxburg, U.~V., Bengio, S., Wallach, H., Fergus, R., Vishwanathan, S., and Garnett, R. (eds.), \emph{Advances in Neural Information Processing Systems}, volume~30. Curran Associates, Inc., 2017.
\newblock URL \url{https://proceedings.neurips.cc/paper_files/paper/2017/file/3f5ee243547dee91fbd053c1c4a845aa-Paper.pdf}.

\bibitem[Wang et~al.(2019)Wang, Zhao, Lioma, Li, Zhang, and Simonsen]{wang2020encoding}
Wang, B., Zhao, D., Lioma, C., Li, Q., Zhang, P., and Simonsen, J.~G.
\newblock Encoding word order in complex embeddings.
\newblock \emph{ArXiv}, abs/1912.12333, 2019.
\newblock URL \url{https://api.semanticscholar.org/CorpusID:209516262}.

\bibitem[Wang et~al.(2020{\natexlab{a}})Wang, Pino, Wu, and Gu]{wang2020covost}
Wang, C., Pino, J., Wu, A., and Gu, J.
\newblock {C}o{V}o{ST}: A diverse multilingual speech-to-text translation corpus.
\newblock In \emph{Proceedings of The 12th Language Resources and Evaluation Conference}, pp.\  4197--4203, Marseille, France, May 2020{\natexlab{a}}. European Language Resources Association.
\newblock ISBN 979-10-95546-34-4.
\newblock URL \url{https://www.aclweb.org/anthology/2020.lrec-1.517}.

\bibitem[Wang et~al.(2020{\natexlab{b}})Wang, Li, Khabsa, Fang, and Ma]{linformer2020}
Wang, S., Li, B.~Z., Khabsa, M., Fang, H., and Ma, H.
\newblock Linformer: Self-attention with linear complexity, 2020{\natexlab{b}}.
\newblock URL \url{https://arxiv.org/abs/2006.04768}.

\bibitem[Wang \& Chen(2020)Wang and Chen]{wang2020position}
Wang, Y.-A. and Chen, Y.-N.
\newblock What do position embeddings learn? an empirical study of pre-trained language model positional encoding.
\newblock In Webber, B., Cohn, T., He, Y., and Liu, Y. (eds.), \emph{Proceedings of the 2020 Conference on Empirical Methods in Natural Language Processing (EMNLP)}, pp.\  6840--6849, Online, November 2020. Association for Computational Linguistics.
\newblock \doi{10.18653/v1/2020.emnlp-main.555}.
\newblock URL \url{https://aclanthology.org/2020.emnlp-main.555}.

\bibitem[Wu et~al.(2020)Wu, Lan, Qian, Gu, Geramifard, and Yu]{memformer}
Wu, Q., Lan, Z., Qian, K., Gu, J., Geramifard, A., and Yu, Z.
\newblock Memformer: A memory-augmented transformer for sequence modeling, 2020.
\newblock URL \url{https://arxiv.org/abs/2010.06891}.

\bibitem[Xiong et~al.(2021)Xiong, Zeng, Chakraborty, Tan, Fung, Li, and Singh]{xiong2021nystromformer}
Xiong, Y., Zeng, Z., Chakraborty, R., Tan, M., Fung, G., Li, Y., and Singh, V.
\newblock Nystr{\"o}mformer: A nystr{\"o}m-based algorithm for approximating self-attention.
\newblock 2021.

\bibitem[Zaheer et~al.(2020)Zaheer, Guruganesh, Dubey, Ainslie, Alberti, Ontanon, Pham, Ravula, Wang, Yang, and Ahmed]{bigbird2020}
Zaheer, M., Guruganesh, G., Dubey, A., Ainslie, J., Alberti, C., Ontanon, S., Pham, P., Ravula, A., Wang, Q., Yang, L., and Ahmed, A.
\newblock Big bird: Transformers for longer sequences.
\newblock 2020.
\newblock \doi{10.48550/ARXIV.2007.14062}.
\newblock URL \url{https://arxiv.org/abs/2007.14062}.

\bibitem[Zhong et~al.(2020)Zhong, Lin, Chen, Li, and Huang]{zhong2019long}
Zhong, G., Lin, X., Chen, K., Li, Q., and Huang, K.
\newblock Long short-term attention.
\newblock In Ren, J., Hussain, A., Zhao, H., Huang, K., Zheng, J., Cai, J., Chen, R., and Xiao, Y. (eds.), \emph{Advances in Brain Inspired Cognitive Systems}, pp.\  45--54, Cham, 2020. Springer International Publishing.
\newblock ISBN 978-3-030-39431-8.

\end{thebibliography}
\bibliographystyle{icml2024}

\appendix

\onecolumn

\newpage

\section{Appendix}
\subsection{Licensing Information}
\label{sec:license}
Fairseq \cite{fairseq2019} is MIT-licensed and widely available for non-commercial use. The Long-Range Arena benchmark \citep{tay2021long} is licensed via Apache 2.0. SimulEval \citep{simuleval2020}, Wikitext-103b \citep{merity2016pointer}, and MuST-C \citep{mustc2021} are licensed via Creative Commons (CC BY-SA 4.0, CC BY-SA 3.0, and CC BY-NC-ND 4.0 respectively).

\subsection{Hardware Details for Training and Evaluation}
\label{sec:hardware}
All models were trained and evaluated on two NVIDIA Tesla V100 32 GB GPUs for LRA and SimulST, except for during evaluation via SimulEval where they operated on a Intel Xeon Platinum 8168 CPU. Autoregressive language models were trained and evaluated on four NVIDIA Tesla V100 32 GB GPUs. 

\subsection{Computational Costs of Experimentation}
\label{sec:comp_cost}
We estimate that results related to the LRA benchmark required approximately 30 GPU hours to gather with perhaps another 60 GPU hours related to experimentation. Concerning SimulST, we estimate that approximately 18 GPU days were required to generate the results with another 4 GPU days related to experimentation.  Autoregressive language modeling required approximately 8 GPU days for experimentation and approximately 60 GPU days to generate the results in this paper. The aforementioned values are normalized for single GPU-usage.

\subsection{Controlling for Representational Capacity in LRA}
\label{sec:repr_cap}
While the Long-Range Arena benchmark \citep{tay2022efficient} requires models to be close in parameter count to one another, so as to hold similar representational capacity, it allows for around a 10\% parameter increase or so if required for the model's functionality. Reformer \citep{reformer2020} and Linformer \citep{linformer2020}, for example, both requre potentially significant parameter increases to properly function. Typically, no additional measures are taken to control for any possible representational capacity increase, but in the interest of ensuring that the LeaP module makes a significant improvement on model performance aside from a simple parameter count increase, we provide some additional results on the LRA benchmark for cosFormer with an average of a 1.05x parameter increase in Table \ref{tab:cosformer_control}.

\begin{table}[h]
    \centering
    \caption{Results on the Long-Range Arena benchmark for a cosFormer implementation with an average increased parameter count of 1.05x.}
    \begin{tabular}{l|ccccc|c}
    \toprule
    Attention Mechanism &  ListOps & Text Cls. & Text Rtr. & Path-32 & Img. Cls. & Avg. \\
    \midrule
    cosformer (1.05x Parameters) & 38.69 & 63.53 & 79.36 & 67.06 & 37.62 & 57.65 \\
    \bottomrule
    \end{tabular}
    \label{tab:cosformer_control}
\end{table}

As observed in Table \ref{tab:cosformer_control}, cosFormer exhibits a minor performance increase for most tasks with the exception of text classification, where improvement is very noticeable. However, scores for ListOps and image classification actually decrease. The latter is not altogether surprising, as the image classification task is structured in a very challenging way in the LRA benchmark (i.e. no ViT-type augmentations \citep{dosovitskiy2021image} are allowed). The above scores result in an RCP$_{mem}$ of 3.78, far below the score of LeaPformers at 5.04 (reusing statistical information from Table \ref{eq:rcp_mem} for ease of comaprison). Overall, it can be concluded that LeaP modules improve performance beyond just an increase in representational capacity. 

\subsection{Vast Efficient Transformer Design Space}
\label{sec:eff_design}
While we cover a comparison of rather seminal works in this paper that we believe are representative of the efficient transformer space, there are a massive amount of efficient transformer designs \citep{tay2022efficient} such that it would be impractical to compare our proposed approach to them all. Just for sparsity-focused transformers alone, there are numerous options with varying degrees of complexity. Sparse Transformer \citep{sparsetransformer}, for example, represents an initial attempt at reducing complexity in this manner alongside its many descendants \citep{image_transf, zhong2019long, longformer2020, etc_ainslie2020, tay2020sparse} eventually culminating in the proposal of BigBird \citep{bigbird2020}, although new sparsity-focused methods are still likely being developed. 

Along the lines of sparse transformer development, other classes of efficient transformers exhibit similar progressions. Regarding softmax approximations with random/sampling-based methods, Random Fourier Attention \citep{rfapeng2021} is an alternative to Performer \citep{performer2020}. Skyformer \citep{chen2021skyformer} was preceded by Nystromformer \citep{xiong2021nystromformer}. TransNormer \citep{qin-etal-2022-devil} is a more recent approach that combines blocked local attention with kernalized linear attention, across two disparate sets of transformer layers (e.g. layers 1-4 might contain one type, layers 5-12 might contain another), although that makes it slightly architecture reliant. 

\subsubsection{Alternative Efficient Architectures: State Space Machines}
A more recently popularized option, the modern State Space Machine (SSM) family of models has emerged as a potential sequence modeling competitor to transformers. In short, SSMs function by substituting explicit recurrent behavior with variants of convolutional kernels, allowing for full parallelizability during training while potentially returning to efficient recurrent behavior during inference (not as applicable for bi-directional sequence modeling). Modern SSMs include the popular S4 family of models \citep{gu2022efficiently} and the  novel Mamba architecture \citep{gu2023mamba}. While this work primarily focuses on exploring and evaluating architectural options for efficient attention mechanisms, we provide some results for S4D, a SSM with a diagonal kernel in the S4 family of architectures, in Table \ref{tab:s4_results}. 

\begin{table}[h]
    \centering
    \caption{Results on the Long-Range Arena benchmark for S4D, a model in the S4 family of SSMs.}
    \begin{tabular}{l|ccccc|c}
    \toprule
    Model &  ListOps & Text Cls. & Text Rtr. & Path-32 & Img. Cls. & Avg. \\
    \midrule
    S4D & 17.79 & 64.67 & 69.81 & 52.34 & 40.86 & 49.09 \\
    \bottomrule
    \end{tabular}
    \label{tab:s4_results}
\end{table}

The results observed in Table \ref{tab:s4_results} are abnormally poor compared linear transformers examined in this work, especially for ListOps and Pathfinder-32. More alarmingly, they contrast harshly with the reported results in S4 publications \citep{gu2022efficiently}, where S4 SSMs tend to perform excellently compared to transformers. There are a few intuitive possible reasons for this observed performance gap. The first is that the Skyformer LRA environment, which we mostly port over, is focused on validating smaller models. It's very possible that the S4 family of SSMs simply doesn't function well with so few parameters. The second is that training was perhaps unstable, as certain S4 parameters are extremely sensitive to learning rate. Our attempts to control for that possibility did not yield meaningful improvements, however. Regardless of the poor prediction quality results, S4D did perform efficiently, yielding up to a 1.7x throughput increase compared to the next fastest linear attention mechanism. This ultimately results in S4D having an RCP$_{mem}$ of around 1.66 (reusing statistical information from Table \ref{tab:rcp_results} to render comparison easy).


\subsection{RoPE with Linear Attention Elaboration}
\label{sec:rope}
While not fully elaborated upon in the paper, we provide data for a single possible RoPE \cite{su2022roformer} linear transformer by augmenting the seminal Linear Transformer \cite{kathrapalous20} with rotary positional embedding. The provided results for this model are based on one with no additional adaptations towards linear transformer functionality beyond what is mentioned in the original publication detailing RoPE. In our tests, additional assurances (e.g. summation to unity in rows of attention matrix, were it to be calculated) did not significantly affect results. 

We touch on the formulation of this linear attention fusion of RoPE and Linear Attention in Equation \ref{eq:rope_reweight}. $R$ represents RoPE's re-weighting function acting as a rotational transform and $\theta$ represents the set of rotation constants defined by head dimensionality $d$. Further details can be found in RoPE's original publication.

\begin{equation}
\begin{aligned}
& \quad \quad \quad \: \sigma(i, j) = R^d_{\theta, j-i} = (R^d_{\theta, i})^TR^d_{\theta, j} \\
& \quad \: \: S(Q_{h,i}, K_{h,j}^T) = S_q(Q_{h,i})\sigma(i, j)S_k(K_{h,j}^T) \\
&S(Q_{h,i}, K_{h,j}^T) = (S_q(Q_{h,i})(R^d_{\theta, i})^T)(R^d_{\theta, j}S_k(K_{h,j}^T)) \\
\end{aligned}
\label{eq:rope_reweight}
\end{equation}

\subsection{LeaPformers SimulST Latency on MuST-C with Elaboration}
\label{sec:simulst_latency}
While not included in the main body of this paper, as our focus was to demonstrate the performance of LeaPformers in terms of translation quality, we provide some results related to SimulST latency in this section. As a qualifier, MuST-C \citep{mustc2021} and SimulST datasets in general tend to be formed of relatively short, single sentence samples. Given the latency profile of LeaPformers and similar linear attention mechanisms, it would be expected that only minor throughput benefits are observed, if throughput benefits are observed at all. Moreover, text as a modality usually favors smaller sequence sizes when engaging in single-to-few sentence translation samples (especially if a model vocabulary is formed of large subwords). Finally, the particular architecture that we employ in this work as a baseline for SimulST downsamples the acoustic features by a factor of 4, reducing the sequence length for the primary, transformer-based body of the model. Regardless of these cavests, SimulST latency results can be observed in Table \ref{tab:simulst_latency} with CA-AL referring to computationally aware average lagging \citep{mma19, simulst_xma20}, the seminal latency/lagging metric of most SimulST works. 

\begin{table}[h]
    \centering
    \caption{Latency results for SimulST experiments on a subset of long-sequence samples in the validation set (approximately 27\% of wall clock run-time). Generation is slightly constrained (minimum token generation steps based on read decisions and maximum set as 10 tokens beyond that minimum).}
    \begin{tabular}{l|cccc}
    \toprule
    Attention Mechanism &  CA-AL (ms) & Comp. Only (ms) & Comp. \% Reduc. & w/o Conv. Pre-net \\
    \midrule
    Softmax & 6841 & 2721 & N/A & N/A \\
    LeaPformer Dec. Self-Attn. & 6864 & 2682 & \hphantom{0}2\% & \hphantom{0}2\% \\
    LeaPformer Dec. Cross-Attn. & 7026 & 2837 & -4\% & -5\% \\
    LeaPformer Enc. Self-Attn. & 6457 & 2276 & 16\% & 21\% \\
    \bottomrule
    \end{tabular}
    \label{tab:simulst_latency}
\end{table}

Not entirely unexpectedly, decoder cross-attention actually performs worse than softmax in our tests. This is due to having to update the cached $K^TV$ intermediate state any time the encoder's acoustic representation is updated, which is unique to simultaneous tasks (we note that this can be avoided somewhat with unidirectional encoders, which we do not employ in our baseline architecture). At long enough sequence lengths, these updates would still result in faster translation, but samples sizes in MuST-C are too small for this behavior to manifest. Contrastingly, decoder self-attention improves slightly and encoder self-attention improves noticeably, resulting in up to a 21\% reduction in the purely computational latency of transformer model components. 

\subsection{Why Not [Insert Alternative Strategy] Instead of LeaP?}
\label{sec:alt_strat}
For the sake of brevity, we could not describe every attempted strategy at adapting some re-weighting functions, specifically the one employed by cosFormer \citep{zhen2022cosformer}, and exhaustively describe why they didn't work in the main body of this paper as there are many obvious options. We will take this opportunity to iterate over a few simple approaches that we attempted to employ or that were touched upon by \citealt{agostinelli2023improving}, whose findings are relevant.

\subsubsection{Why Not Step the Sequence Length?}
The most obvious adaptation option for autoregressive, and even simultaneous, tasks would be to just increment the sequence length that re-weighting functions like the one employed by cosFormer are dependent upon for every decoding time-step (and encoding time-step, for simultaneous applications). This was explored early on and resulted in extreme inference quality degradation for en-de SimulST results (BLEU of around 1.1), and thinking about the resulting re-weighting function across time-steps makes it easy to intuit why that might be. 

In the case of cosFormer's re-weighting function, doing this results in static query proportions at every time-step (slightly different behavior for chunked simultaneous encoders). While this can be fine for decoder self-attention (although it did not work well in our tests), it is essentially impossible to apply to decoder cross-attention in a way that meaningfully attenuates the query and key matrices because, according to the stepping re-weighting function, every query is "placed" identically in the sequence.

\begin{figure*}[t]
    \centering
    \includegraphics[scale=0.34]{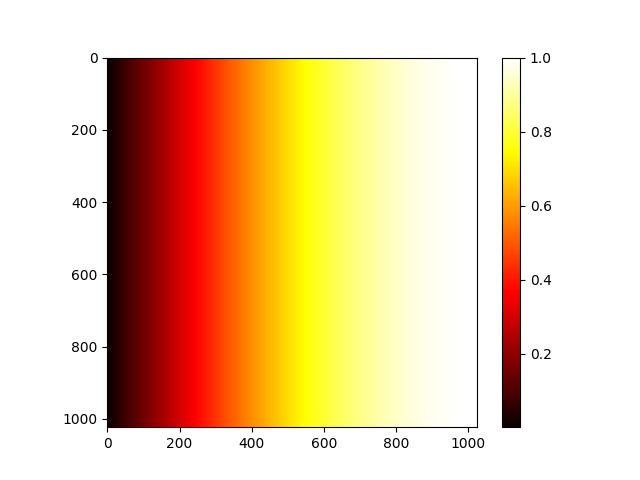}
    \includegraphics[scale=0.34]{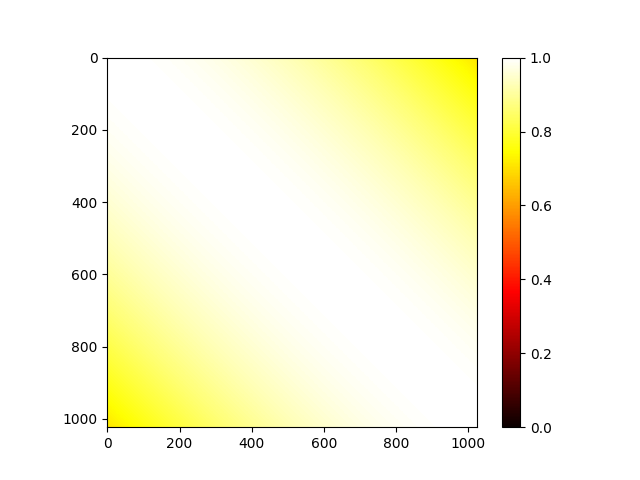}
    \includegraphics[scale=0.34]{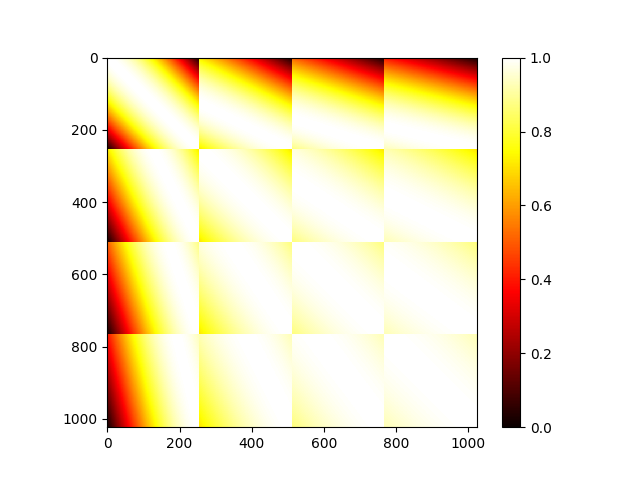}
    \caption{Re-weighting heatmaps of decoder cross-attention during various alternative strategies to LeaP: stepping sequence length (left), max sequence length (middle), stepping the max sequence length intermittently (right).}
    \label{fig:multiple_strategies}
\end{figure*}

\subsubsection{Why Not Set a Max Sequence Length?}
An understandable approach would be to simply set a large, maximum sequence length. In our tests, this worked somewhat adequately, but suffered from unused expressivity that resulted in noticeable inference quality degradation. To ensure positivity for all samples, a very large maximum sequence length needs to be chosen for autoregressive tasks, which means that for samples that are extremely small, queries and keys are effectively barely attenuated.

\subsubsection{Why Not a Stepping Max Length?}
As a subset of the above stepping and maximum length strategies, one could attempt to set a maximum sequence length that steps only occasionally. For example, an autoregressive sample whose reference is of length $N$ (unknown during inference, average training set reference is $\frac{N}{4}$) might have a hypothetical stepping maximum sequence length of around $\frac{N}{4}$ that would step by $\frac{N}{4}$ upon the generated sequence growing. This should, intuitively, resolve problems with naive stepping and lost expressivity for maximum sequence lengths so long as the stepping maximum is reasonably small. While more effective than either of the above strategies, generative behavior around borders for incrementation (i.e. nearing limit of current sequence length) tended to result in unusual model behavior that lead to reduced inference quality.

\subsubsection{Why Not Predict Sequence Length with Neural Networks or Statistics?}
While \citealt{agostinelli2023improving} propose this exact strategy, and find that it works reasonably well in practice for general, non-fixed length autoregressive tasks, such methods are almost always impossible to apply for simultaneous tasks with low-lagging schedules between the source context and target sequence. This is because, usually, complete or almost complete source context is required to make accurate predictions. For example, asking a compact neural network to predict the target sequence length after taking in 400 ms of audio of a 5000 ms audio clip is clearly an exceedingly difficult task unless there are clear indicators of sequence length (e.g. audio prefaced with total audio length, doesn't work for most practical simultaneous tasks).

\subsection{Elaboration on Throughput Observations for Efficient Attention Mechanisms}
\label{sec:thrpt}
It is worth lightly elaborating upon the throughput observations provided in Table \ref{tab:lra_results_latency}, as some of them may be surprising to readers. For example, it is initially unclear as to why truly linear attention mechanisms (e.g. Linear Attention from \citealt{kathrapalous20}) are exceedingly fast in comparison to some alternatives, like BigBird \citep{bigbird2020}, that rely on sparsity-focused approaches. While sparsity-focused approaches can be theoretically fast, inefficiencies in memory accesses can result in difficult to avoid slowdowns, especially on general purpose hardware. Indeed, the original publication of BigBird acknowledges this and recommends a block-based approach to its various components to avoid memory access bottlenecks, which has been implemented for the version of BigBird that we test in this paper and is implemented for every version of BigBird that we are aware of (i.e. present in the original LRA publication).

Despite employing a block-based accessing approach for BigBird, findings across efficient attention mechanism literature continue to find that, generally, this particular efficient attention scheme scales poorly with respect to throughput for extremely long sequences \citep{tay2021long, chen2021skyformer, zhen2022cosformer}, although this depends in part on hyperparameters that are employed to determine how such sparsity-focused methods behaved (i.e. how often they actually compute information). Other efficient attention mechanisms, like Reformer \citep{reformer2020} and Linformer \citep{linformer2020}, can also be considered somewhat tuning-heavy to find a desirable performance-to-throughput trade-off versus truly linear attention mechanisms. Given that we do not explore tweaking hyperparameters in this paper to boost the performance of any architecture, we consider it out of the scope of this work to search for anything beyond the hyperparameters employed by Skyformer \citep{chen2021skyformer}, whose PyTorch implementation of the LRA is what our implementation is based upon.

\subsubsection{Brief BigBird Scaling Example}
For BigBird in particular, we can consider a 4k sequence example and employ the relevant hyperparameters provided in the next section, those being a block size of 64 on each side and 3 random blocks, 3 local blocks, and 2 global blocks. Considering just random attention alone, this results in approximately $2 \times 192N_1d$ FLOPs per attention head, where $N_1$ is the query length and is 4k in this case and $d$ is head dimensionality set to 32 in this case, if BigBird attention is computed optimally, ignoring softmax normalization and projections. Compared to Linear Attention, approximately $N_2d^2 + N_1d^2$ FLOPs are required, where $N_2$ is the key and value length and is 4k in this case. This ends up resulting in an approximate 6x increase in FLOPs from Linear Attention to BigBird, underscoring clearly BigBird's scaling issues. While one could reduce the parameters chosen for BigBird in this paper to increase throughput, that would likely correspond to a reduction in inference quality.

\newpage

\subsection{Model Architectures and Hyperparameters}
\label{sec:arch}
Below, we list all architectural details and relevant training hyperparameters to reproduce our experiments. Aside from models explicitly including RoPE in our tests, all other models employed absolute positional encoding (APE). 

Regarding specific attention mechanism hyperparameters: Linformer \citep{linformer2020} employs a $k$ low-rank factor of 2, Reformer \citep{reformer2020} employs two hashes for its LSH method, Performer \citep{performer2020} models an exponential kernel with 128 rows for the orthogonal random matrices, BigBird \citep{bigbird2020} employs blocks of size 64 on each side and chooses 3 blocks for random attention (e.g. 192 tokens per query in the 4k sequence case), 3 blocks for localized attention (e.g. 192 tokens per query), and 2 blocks for global attention (e.g. 4k tokens for the first 128 queries, then 128 tokens per query), and Skyformer \citep{chen2021skyformer} employs a sampling factor of 4. All of these hyperparameters are identical to those employed in the original Skyformer publication.


\subsubsection{LRA: ListOps}
Below are the architectural details for our ListOps models on the LRA benchmark:
\begin{itemize}
    \item Encoder Layers: 2
    \item Transformer Dim. $d_{model}$: 64
    \item Attention Heads: 2
    \item FFN Hidden Dim. $d_{ffn}$: 128
    \item LeaP Downsample Factor: 1
\end{itemize}

The models for LRA ListOps, were optimized with Adam with classical parameters. The models were trained with batches of size 32, warmed up for 1000 updates and linearly climbing to a learning rate of 1e-4. A linear learning rate decay was employed with 20000 updates in total. A CLS token was used for classification. Dropouts of 0.1 were employed when applicable. 


\subsubsection{LRA: Pathfinder-32}
Below are the architectural details for our Pathfinder-32 models on the LRA benchmark:
\begin{itemize}
    \item Encoder Layers: 2
    \item Transformer Dim. $d_{model}$: 64
    \item Attention Heads: 2
    \item FFN Hidden Dim. $d_{ffn}$: 128
    \item LeaP Downsample Factor: 1
\end{itemize}

The models for LRA Pathfinder-32, were optimized with Adam with classical parameters. The models were trained with batches of size 128, warmed up for 300 updates and linearly climbing to a learning rate of 2e-4. A linear learning rate decay was employed with 50000 updates in total. A CLS token was used for classification. Dropouts of 0.1 were employed when applicable. 

\newpage

\subsubsection{LRA: Text Retrieval}
Below are the architectural details for our Text Retrieval models on the LRA benchmark:
\begin{itemize}
    \item Encoder Layers: 2
    \item Transformer Dim. $d_{model}$: 64
    \item Attention Heads: 2
    \item FFN Hidden Dim. $d_{ffn}$: 128
    \item LeaP Downsample Factor: 2
\end{itemize}

The models for LRA Text Retrieval, were optimized with Adam with classical parameters. The models were trained with batches of size 16, warmed up for 800 updates and linearly climbing to a learning rate of 2e-4. A linear learning rate decay was employed with 50000 updates in total. A CLS token was used for classification. Dropouts of 0.1 were employed when applicable. 


\subsubsection{LRA: Text Classification}
Below are the architectural details for our Text Classification models on the LRA benchmark:
\begin{itemize}
    \item Encoder Layers: 2
    \item Transformer Dim. $d_{model}$: 64
    \item Attention Heads: 2
    \item FFN Hidden Dim. $d_{ffn}$: 128
    \item LeaP Downsample Factor: 2
\end{itemize}

The models for LRA Text Classification, were optimized with Adam with classical parameters. The models were trained with batches of size 32, warmed up for 100 updates and linearly climbing to a learning rate of 2e-4. A linear learning rate decay was employed with 20000 updates in total. A CLS token was used for classification. Dropouts of 0.1 were employed when applicable. 

\subsubsection{LRA: Image Classification}
Below are the architectural details for our Image Classification models on the LRA benchmark:
\begin{itemize}
    \item Encoder Layers: 2
    \item Transformer Dim. $d_{model}$: 64
    \item Attention Heads: 2
    \item FFN Hidden Dim. $d_{ffn}$: 128
    \item LeaP Downsample Factor: 1
\end{itemize}

The models for LRA Image Retrieval, were optimized with Adam with classical parameters. The models were trained with batches of size 256, warmed up for 200 updates and linearly climbing to a learning rate of 1e-4. A linear learning rate decay was employed with 30000 updates in total. A CLS token was used for classification. Dropouts of 0.1 were employed when applicable. 


\subsubsection{Autoregressive Language Modeling Models}
Below are the architectural details for our autoregressive language modeling models, identical to \citealt{baevski2019adaptive}'s implementation

\begin{itemize}
    \item Decoder Layers: 16
    \item Transformer Dim. $d_{model}$: 512
    \item Attention Heads: 8
    \item FFN Hidden Dim. $d_{ffn}$: 2048
    \item Adaptive Input/Softmax Cutoffs: 20k, 60k
    \item LeaP Downsample Factor: 4
\end{itemize}

The models for autoregressive language modeling were optimized via Nesterov's accelerated gradient method with a momentum of 0.99 with a gradient renormalization threshold of 0.1. The models were trained with an initial learning rate of 1e-7 scaling to 1.0 over 8000 updates. From there, we define a cosine-based decaying schedule with periods of 140k updates and train for 150k updates with a minimum learning rate of 1e-9. Dropouts of 0.3 were used for linear layers with dropouts of 0.1 used elsewhere when applicable.


\subsubsection{SimulST Models}
Below are the architectural details for our SimulST models:
\begin{itemize}
    \item Encoder Layers: 12
    \item Decoder Layers: 6
    \item Transformer Dim. $d_{model}$: 256
    \item Attention Heads: 8
    \item FFN Hidden Dim. $d_{ffn}$: 1024
    \item Conv. Pre-net Layers: 2
    \item Conv. Pre-net Kernel Size: 3
    \item Conv. Pre-net Stride: 2
    \item LeaP Downsample Factor: 4
\end{itemize}

The models for SimulST tasks were optimized via Adam with classical parameters and a learning rate set to 6e-4 with an identical learning rate scheduler. The models were trained with dynamic batching, warmed up for 6000 updates, starting with a learning rate of 1e-4, and trained for around 18000 updates in total with gradients clipped to 10.0. Dropouts of 0.1 were used for all linear layers and attention. SimulST models were trained with a wait-k of 5 and pre-decision ratio of 9. 

\end{document}